\documentclass{article}

\usepackage{microtype}
\usepackage{graphicx}
\usepackage{booktabs} 
\usepackage{amsmath}
\usepackage{amsfonts}
\usepackage{amsthm}
\usepackage{bm}
\usepackage{subcaption}
\usepackage{makecell}
\usepackage{multirow}
\usepackage{booktabs} 

\usepackage[]{hyperref}

\usepackage[accepted]{icml2020}

\newcommand{\cbar}{\,|\,}
\newcommand{\X}{\mathbf{X}}

\newcommand{\Z}{\mathbf{Z}}
\newcommand{\x}{\mathbf{x}}

\newcommand{\PC}{\mathcal{P}}
\newcommand{\graph}{\mathcal{G}}

\newcommand{\scope}{\psi}
\newcommand{\pa}{\mathbf{pa}}
\newcommand{\ch}{\mathbf{ch}}
\newcommand{\val}{\mathbf{val}}

\newcommand{\node}{\mathsf{N}}
\newcommand{\sumnode}{\mathsf{S}}
\newcommand{\prodnode}{\mathsf{P}}
\newcommand{\leaf}{\mathsf{L}}

\newcommand{\w}{w}

\newtheorem{definition}{Definition}

\icmltitlerunning{Einsum Networks: Fast and Scalable Learning of Tractable Probabilistic Circuits}

\begin{document}

\twocolumn[
\icmltitle{Einsum Networks: Fast and Scalable Learning of \\ Tractable Probabilistic Circuits}



\icmlsetsymbol{equal}{*}

\begin{icmlauthorlist}
\icmlauthor{Robert Peharz}{tue}
\icmlauthor{Steven Lang}{tuds}
\icmlauthor{Antonio Vergari}{ucla}
\icmlauthor{Karl Stelzner}{tuds}
\icmlauthor{Alejandro Molina}{tuds}
\icmlauthor{Martin Trapp }{tug}
\icmlauthor{Guy Van den Broeck}{ucla}
\icmlauthor{Kristian Kersting}{tuds}
\icmlauthor{Zoubin Ghahramani}{cam}
\end{icmlauthorlist}

\icmlaffiliation{tue}{Eindhoven University of Technology}
\icmlaffiliation{tuds}{Technical University Darmstadt}
\icmlaffiliation{tug}{Graz University of Technology}
\icmlaffiliation{ucla}{University California Los Angeles}
\icmlaffiliation{cam}{University of Cambridge}

\icmlcorrespondingauthor{Robert Peharz}{r.peharz@tue.nl}
\icmlcorrespondingauthor{Steven Lang}{steven.lang.mz@gmail.com}

\icmlkeywords{Tractable Probabilistic Models}

\vskip 0.3in
]



\printAffiliationsAndNotice{}  

\begin{abstract}
Probabilistic circuits (PCs) are a promising avenue for probabilistic modeling, as they permit a wide range of exact and efficient inference routines.
Recent ``deep-learning-style'' implementations of PCs strive for a better scalability, but are still difficult to train on real-world data, due to their sparsely connected computational graphs.
In this paper, we propose Einsum Networks (EiNets), a novel implementation design for PCs, improving prior art in several regards.
At their core, EiNets combine a large number of arithmetic operations in a single monolithic einsum-operation, leading to speedups and memory savings of up to two orders of magnitude, in comparison to previous implementations.
As an algorithmic contribution, we show that the implementation of Expectation-Maximization (EM) can be simplified for PCs, by leveraging automatic differentiation.
Furthermore, we demonstrate that EiNets scale well to datasets which were previously out of reach, such as SVHN and CelebA, and that they can be used as faithful generative image models.
\end{abstract}

\section{Introduction}   \label{sec:introduction}

The central goal of probabilistic modeling is to approximate the data-generating distribution, in order to answer statistical queries by means of inference.
In recent years many novel probabilistic models based on deep neural networks have been proposed, such as Variational Autoencoders (VAEs) \cite{Kingma2014} (formerly known as Density Networks \cite{Mackay1995}), Normalizing Flows (Flows) \cite{Rezende2015,Papamakarios2019}, Autoregressive Models (ARMs) \cite{Larochelle2011,Uria2016}, and Generative Adversarial Networks (GANs) \cite{Goodfellow2014}.
While all these models have achieved impressive results on large-scale datasets, i.e.~they have been successful in terms of \textbf{representational power} and \textbf{learning}, they unfortunately fall short in terms of \textbf{inference}, a main aspect of probabilistic modeling and reasoning \cite{Pearl1988,Koller2009}.
All of the mentioned models allow to draw unbiased samples, enabling inference via Monte Carlo estimation.
This strategy, however, becomes quickly unreliable and computational expensive for all but the simplest inference queries.
Also other approximate inference techniques, e.g.~variational inference, are often biased  and their inference quality might be hard to analyse.
Besides sampling, only ARMs and Flows support efficient evaluation of the probability density for a given sample, 
which can be used, e.g., for model comparison and outlier detection.

However, even for ARMs and Flows the following inference task is computationally hard:
Consider a density $p(X_1, \dots, X_N)$ over $N$ random variables, where $N$ might be just in the order of a few dozens.
For example, $X_1, \dots, X_N$ might represent medical measurements of a particular person drawn from the general population modeled by $p(X_1, \dots, X_N)$.
Now assume that we split the variables into three disjoint sets $\X_q$, $\X_m$, and $\X_e$ of roughly the same size, and that we wish to compute
\begin{equation}   \label{eq:conditional_density}
p(\x_q \cbar \x_e) = 
\frac{\int p(\x_q, \x'_m, \x_e) \mathrm{d}\x'_m}
{\int \int p(\x'_q, \x'_m, \x_e) \mathrm{d}\x'_q \mathrm{d}\x'_m},
\end{equation}
for arbitrary values $\x_q$ and $\x_e$. 
In words, we wish to predict query variables $\X_q$, based on evidence $\X_e = \x_e$, while accounting for (marginalizing) all possible values of missing variables $\X_m$.
Conditional densities like \eqref{eq:conditional_density} highlight the role of generative models as ``multi-purpose predictors'', since the choice of $\X_q$, $\X_m$, and $\X_e$ can be arbitrary.
However, evaluating conditional densities of this form is notoriously hard for any of the models above, which represents a principal drawback of these methods.

These shortcomings in terms of inference have motivated a growing stream of research on \emph{tractable probabilistic modeling}, i.e.~to construct a constrained class of probabilistic models, in which inference queries like \eqref{eq:conditional_density} can be computed \emph{exactly} and \emph{efficiently} (in polynomial time).
One of the most prominent families of tractable models are \emph{probabilistic circuits} (PCs),\footnote{We adopt the name \emph{probabilistic circuits}, as suggested by \cite{VanDenBroeck2019}, which serves as an umbrella term for many structurally related models, like arithmetic circuits, sum-product networks, cutset networks, etc. See Section~\ref{sec:probabilistic_circuits} for details.}
which represent probability densities via a computational graph of i) \emph{mixtures} (convex sum nodes), ii) \emph{factorizations} (product nodes), and iii) \emph{tractable distributions} (leaves, input nodes).
A key structural property of PCs is \emph{decomposability} \cite{Darwiche2003}, which ensures that \emph{any} integral, like those appearing in \eqref{eq:conditional_density}, can be computed in \emph{linear} time of the circuit size.
PCs can be equipped with further structural constraints, which unlock a wider range of tractable inference routines -- see Section~\ref{sec:probabilistic_circuits} for an overview.
These structural constraints, however, make it hard to work with PCs in practice, as they lead to highly sparse and cluttered computational graphs, which are inapt for current machine learning frameworks.
Furthermore, PCs are typically implemented in the log-domain, which slows down learning and inference even further.

In this paper, we propose a novel implementation design for PCs, which ameliorates these practical difficulties, and allows to evaluate and train PCs of up to \emph{two orders of magnitude faster} than previous implementations \cite{Pronobis2017,Peharz2019}.
The central idea is to compute all product and sum operations on the same topological layer using a single monolithic \emph{einsum} operation.\footnote{The \emph{einsum} operation implements the Einstein notation of tensor-product contraction, and unifies standard linear algebra operations like dot product, outer product, and matrix multiplication.}
In that way, the main computational work is lifted by a parallel operation for which efficient implementations, both for CPU and GPU, are readily available in most numerical frameworks.
In order to ensure numerical stability, we extend the well-known log-sum-exp-trick to our setting, leading to the ``log-\emph{einsum}-exp'' trick.
Since our model implements PCs via a hierarchy of large einsum layers, we call our model \emph{Einsum Network} (EiNet).

As further contributions, we present two algorithmic improvements for training PCs.
First, we show that Expectation-Maximization (EM), the canonical maximum-likelihood learning algorithm for PCs \cite{Peharz2017}, can be easily implemented using the gradient of the model's log-probability.
Thus, EM can be implemented using automatic differentiation, readily provided by most machine learning frameworks.
Second, we leverage stochastic online EM \cite{Sato1999} in order to further improve the learning speed of EiNets, or, enable EM for large datasets at all.
In experiments we demonstrate that EiNets can rapidly learn generative models for street view house numbers (SVHN) and CelebA. 
To the best of our knowledge, this is the first time that PCs have been successfully trained on datasets of this size.
EiNets are capable of producing high-quality image samples, while maintaining tractable inference, e.g.~for conditional densities \eqref{eq:conditional_density}.
This can be exploited for arbitrary image inpainting tasks and other advanced inference tasks.

Concerning notation, we use $X$, possibly with sub-scripts, to denote random variables and $x$ to denote a value of $X$.
Sets (or tuples) of random variables are denoted as $\X$, also possibly with sub-script, and their values as $\x$.
We denote the set of possible values for (sets of) random variables with $\val(\cdot)$.
For directed graphs, we use $\pa(\node)$ and $\ch(\node)$ to denote the parents  and children of node $\node$, respectively.

\section{Probabilistic Circuits}   \label{sec:probabilistic_circuits}

Probabilistic circuits (PCs) are a family of probabilistic models which allow a wide range of exact and efficient inference routines.
The earliest representatives of PCs are \emph{arithmetic circuits} (ACs) \cite{Darwiche2002, Darwiche2003}, which are based on \emph{decomposable negation normal forms} (DNNFs) \cite{Darwiche1999, Darwiche2001}, a tractable representation for propositional logic formulas.
Further members of the PC family are \emph{sum-product networks} (SPNs) \cite{Poon2011}, \emph{cutset networks} (CNs) \cite{Rahman2014}, and \emph{probabilistic sentential decision diagrams} (PSDDs) \cite{Kisa2014}.
The main differences between different types of PCs are i) the set of assumed \emph{structural constraints} and tractable inference routines, ii) syntax and representation, and iii) application scenarios.
In order to treat these models in a unified fashion, we adopt the umbrella term \emph{probabilistic circuits} suggested by \cite{VanDenBroeck2019}, and discriminate various instantiates of PCs mainly via their structural properties.

\begin{definition}[Probabilistic Circuit]   
\label{def:probabilistic_circuit}
Given a set of random variables $\X$, a \emph{probabilistic circuit} (PC) $\PC$ is a tuple $(\graph, \scope)$, where $\graph$, denoted as \emph{computational graph}, is a directed acyclic graph (DAG) $(V, E)$ and $\scope \colon V \mapsto 2^{\X}$, denoted as \emph{scope function}, is a function assigning a \emph{scope} to each node in $V$, i.e.~a sub-set of $\X$. 
For internal nodes of $\graph$, i.e.~any node $\node \in V$ which has children, the scope function satisfies $\scope(\node) = \cup_{\node' \in \ch(\node)} \scope(\node')$. 
A \emph{leaf} of $\graph$ computes a probability density
over its scope $\scope(\leaf)$.
All internal nodes of $\graph$ are either \emph{sum nodes} ($\sumnode$) or \emph{product nodes} ($\prodnode$).
A sum node $\sumnode$ computes a convex combination of its children, i.e.~$\sumnode = \sum_{\node \in \ch(\node)} w_{\sumnode,\node} \, \node$, where $\sum_{\node \in \ch(\node)} w_{\sumnode,\node} = 1$, and $\forall \node \in \ch(\sumnode)
\colon w_{\sumnode,\node} \geq 0$.
A product node $\prodnode$ computes a product of its children, i.e.~$\prodnode = \prod_{\node \in \ch(\node)} \node$.
\end{definition}

PCs can be seen as a special kind of neural network, where the first layer computes non-linear functions (probability densities) over sub-sets of $\X$, and all internal nodes compute either weighted sums (linear functions) or products (multiplicative interactions) of their inputs.
Each node $\node$ in a PC sees only a sub-set of the inputs, namely variables in its scope $\scope(\node)$. 
The output of a PC is the value of one or more selected nodes in the computational graph $\graph$.
Typically, one constructs PCs such that they have a single root $\node_{root}$ (node without parents), for which we assume full scope $\scope(\node_{root}) = \X$, and define a density $\PC(\x)$ over $\X$ proportional to $\node_{root}(\x)$, i.e.~$\PC(\x) := \frac{\node_{root}(\x)}{\int_{\val(\X)} \node_{root}(\x) \mathrm{d}\x}$.
PCs as defined above define a density, but do not permit tractable inference yet.
In particular, the normalization constant is hard to compute. 
Various tractable inference routines ``can be bought'' by imposing structural constraints on the PC, which we review next.

\textbf{Decomposability.} 
A PC is \emph{decomposable}, if for each product node $\prodnode \in V$ it holds that $\scope(\node) \cap \scope(\node') = \emptyset$, for $\node, \node' \in \ch(\prodnode)$, $\node \not= \node'$, i.e.~a PC is decomposable if children of product nodes have pairwise non-overlapping scope.
The most important consequence of decomposability is that integrals which can be written as nested single-dimensional integrals---in particular the normalization constant---can be computed in \emph{linear} time of the circuit size \cite{Peharz2015b}.
This follows from the fact that any single-dimensional integral i) commutes with a sum-node and ii) affects only a single child of a product-node.
Consequently, all integrals can be distributed to the PC's leaves, i.e.~we simply need to perform integration at the leaves (which we assume to be tractable), and evaluate the internal nodes of the PC as usual, in a single feedforward pass.

\textbf{Smoothness.}
A PC is \emph{smooth}, if for each sum node $\sumnode \in V$ it holds that $\scope(\node) = \scope(\node')$, for $\node, \node' \in \ch(\sumnode)$, i.e.~a PC is smooth if children of sum nodes have identical scope.
Smoothness does not have particular computational advantages, but leads to a well-defined probabilistic interpretation of PCs.
In particular, in smooth (and decomposable) PCs any node is already a properly normalized density, since i) leaves are densities by assumption, ii) sum nodes are simply \emph{mixture densities}, with their children being the mixture components, and iii) product nodes are factorized densities.
Smooth and decomposable PCs admit a natural interpretation as latent variable models \cite{Zhao2015,Peharz2017}, which admits a natural sampling procedure via ancestral sampling and maximum-likelihood learning via EM.
Smooth and decomposable PCs are often referred to as sum-product networks (SPNs) \cite{Poon2011}.

In this paper, we consider only smooth and decomposable PCs (aka SPNs), which facilitate efficient marginalization and conditioning.
A further structural property, not considered in this paper, is \emph{determinism} \cite{Darwiche2003} which allows for exact probability maximization (\emph{most-probable-explanation}), which is NP-hard in non-deterministic PCs \cite{Campos2011,Peharz2017}.
Moreover, PCs can be equipped with \emph{structured decomposability} \cite{Kisa2014}, a stronger notion of decomposability which allows circuit multiplication and computing certain expectations \cite{Shen2016,Khosravi2019}.

\section{Einsum Networks}   \label{sec:einsum_networks}

\subsection{Vectorizing Probabilistic Circuits}   \label{sec:vectorizing}

An immediate way to yield a denser and thus more efficient layout for PCs is to \emph{vectorize} them.
To this end, we re-define a leaf $\leaf$ to be a \emph{vector} of $K$ densities over $\scope(\leaf)$, rather than a single density.
For example, a leaf computing a Gaussian density is replaced by a vector $[\mathcal{N}(\cdot | \bm{\theta}_1), \dots, \mathcal{N}(\cdot | \bm{\theta}_K)]^T$, each $\mathcal{N}(\cdot | \bm{\theta}_k)$ being a Gaussian over $\scope(\leaf)$, equipped with private parameters $\bm{\theta}_k$.
A product node is re-defined to be an \emph{outer product} $\otimes_{\node \in \ch(\prodnode)} \node$, containing the products of all possible combinations of densities coming from the child vectors.
Since the number of elements in outer products grows exponentially, we restrict the number of children to two (this constraint is frequently imposed on PCs for simplicity).
Finally, we re-define sum nodes to be a vector of $K$ weighted sums, where each individual sum operation has its private weights and computes a convex combination over all the densities computed by its children.
The vectorized version of PCs is frequently called a \emph{region graph} \cite{Dennis2012,Trapp2019} and has been used in previous GPU-supporting implementations \cite{Pronobis2017,Peharz2019}.
It is easily verified, that our desired structural properties---smoothness and decomposability---carry over to vectorized PCs.

For the remainder of the paper, we use symbols $\sumnode$, $\prodnode$, $\leaf$ for the vectorized versions, and refer to them as sum nodes, product nodes, and leaf nodes, respectively, or also simply as sums, products and leaves.
To any single entry in these vectors we explicitly refer to as \emph{entry}, or \emph{operation}.
In principle, the number of entries $K$ could be different for each leaf or sum, which would, however, lead to a less homogeneous PC design.
Therefore, in this paper, we assume for simplicity the same $K$ for all leaves and sums.
Furthermore, we make some simplifying assumptions about the structure of $\graph$.
First, we assume a structure alternating between sums/leaves and products, i.e.~children of sums can only be products, and children of products can only be sums or leaves.
Second, we also assume the root of the PC is a sum node.
These assumptions are commonly made in PC literature and are no restriction of generality.
Furthermore, we also assume that each product node has at most one parent (which must be a sum due to alternating structure).
This is also not a severe structural assumption: If a product has two or more sum nodes as parents, then, by smoothness, these sum nodes have all the same scope.
Consequently, they could be simply concatenated to a \emph{single} sum vector.
Of course, since in this paper we assume a constant length $K$ for all sum and leaf vectors, this assumption requires a large enough $K$.

\subsection{The Basic Einsum Operation}   \label{sec:basic_einsum}

\begin{figure}
\centering
\includegraphics[width=0.35\textwidth]{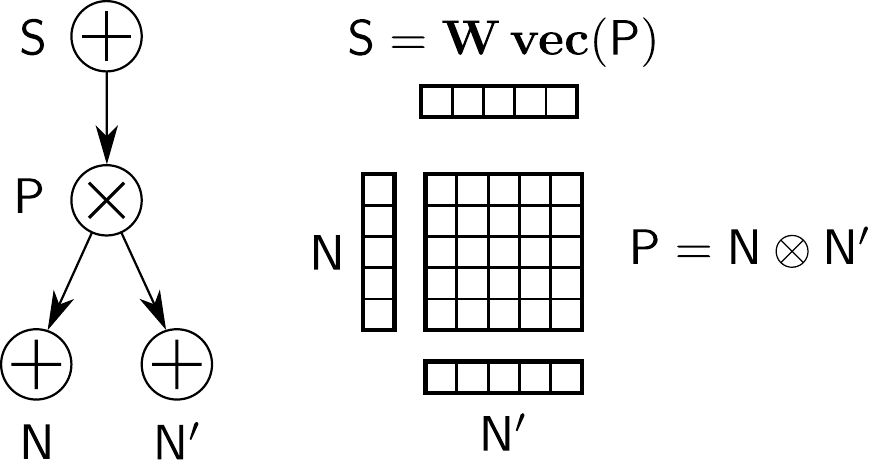}%
\caption{Basic einsum operation in EiNets: A sum node $\sumnode$, with a single child $\prodnode$, which itself has 2 children.
All nodes are vectorized, as described in Section~\ref{sec:vectorizing}, and here illustrated for $K=5$.
}
\label{fig:basic_einsum}
\end{figure}

The core computational unit in EiNets is the vectorized PC excerpt in Fig.~\ref{fig:basic_einsum}, showing a sum node $\sumnode$ with a single product child $\prodnode$, which itself has two children $\node$ and $\node'$ (shown here as sum nodes, but they could also be leaves).
Nodes $\node$ and $\node'$ compute each a vector of $K$ densities, the product node $\prodnode$ computes the outer product of $\node$ and $\node'$, and the sum node $\sumnode$ computes a matrix-vector product $\mathbf{W}\,\mathbf{vec}(\prodnode)$.
Here, $\mathbf{W}$ is an element-wise non-negative $K \times K^2$ matrix, whose rows sum to one, and $\mathbf{vec}(\prodnode)$ unrolls $\prodnode$ to a vector of $K^2$ elements.

Previous PC implementations \cite{Pronobis2017,Peharz2019}, are also based on this core computational unit.
However, for numerical stability, they use a computational workaround in the log-domain:
The outer product is transformed into an ``outer sum'' of log-densities (realized with broadcasting), the matrix multiplication is implemented using a broadcasted sum of $\log \mathbf{W}$ and $\mathbf{vec}(\log \prodnode)$, to which then a log-sum-exp operation is applied, yielding $\log \sumnode$.
This workaround introduces significant overhead and needs to allocate the products explicitly.
Mathematically, however, the PC excerpt in Fig.~\ref{fig:basic_einsum} is a simple multi-linear form, naturally expressed in \emph{Einstein notation}:
\begin{equation}   \label{eq:basic_einsum}
\sumnode_k = \mathbf{W}_{kij} \node_i \node'_{j}.
\end{equation}
Here we have re-shaped $\mathbf{W}$ into a $K \times \ K \times K$ element-wise non-negative tensor, normalized over its last two dimensions, i.e.~$W_{kij} \geq 0$, $\sum_{i,j} W_{kij} = 1$.
The signature in \eqref{eq:basic_einsum} mentions three indices $i$, $j$ and $k$ labeling the axes of $\node$, $\node'$, and $\mathbf{W}$. 
Axes with the same index get multiplied.
Furthermore, any indices not mentioned on the left hand side get summed out.
General-purpose Einstein summations are readily implemented in most numerical frameworks, and usually denoted as \emph{einsum} operation.

However, applying \eqref{eq:basic_einsum} in a naive way would quickly lead to numerical underflow and unstable training.
In order to ensure numerical stability, we develop a technique similar to the classical ``log-sum-exp''-trick.
We keep all probabilistic values in the log-domain, but the weight-tensor $\mathbf{W}$ is kept in the \emph{linear} domain.
Consequently, we need a numerically stable computation for 
\begin{equation}   \label{eq:basic_log_einsum_exp}
\log \sumnode_k = \log \sum_{i,j} W_{kij} \exp(\log\node_i) \exp(\log\node'_{j}).
\end{equation}
Let us define $a = \max_i \log \node_i$ and  $a' = \max_j \log \node'_j$.
Then, we can show that $\log \sumnode_k$ can be computed as
\begin{equation}   \label{eq:basic_log_einsum_exp_trick}
a+a'+\log \sum_{i,j} W_{kij} \exp(\log\node_i - a) \exp(\log\node'_{j} - a').
\end{equation}
To see that \eqref{eq:basic_log_einsum_exp_trick} is correct, note that for the last term we have
\begin{align*}
& \log \sum_{i,j} W_{kij} \exp(\log\node_i - a) \exp(\log\node'_{j} - a') \\
= & \log \sum_{i,j} W_{kij} \exp(- a -a') \exp(\log\node_i) \exp(\log\node'_{j}) \\
= & - a -a' + \log \sum_{i,j} W_{kij} \exp(\log\node_i) \exp(\log\node'_{j}).
\label{eq:basic_log_einsum_exp_trick_reform_term}
\end{align*}
Substituting the last line into \eqref{eq:basic_log_einsum_exp_trick} yields \eqref{eq:basic_log_einsum_exp}, so our log-einsum-exp trick delivers the correct result.
A sufficient condition for numerical stability of \eqref{eq:basic_log_einsum_exp_trick} is that all sum-weights ${W}_{kij}$ are larger than $0$, since in this case the maximal values in vectors $\exp(\log\node - a)$ and $\exp(\log\node' - a')$ are guaranteed to be $1$, leading to a positive argument for the $\log$.
This is not a severe requirement, as positive sum-weights are commonly enforced in PCs, e.g.~by using Laplace smoothing or imposing a positive lower bound on the weights.

Given two $K$-dimensional vectors $\node$, $\node'$ and the $K\times K \times K$ weight-tensor $\mathbf{W}$, our basic einsum operation  \eqref{eq:basic_log_einsum_exp_trick} requires $2K$ exp-operations, $K$ $\log$-operations, $\mathcal{O}(K^3)$ multiplications and $\mathcal{O}(K^3)$ additions. 
We need to store $3K$ values for $\node$, $\node'$, $\sumnode$, while the product operations are not stored explicitly.
In contrast, the indirect implementations of the same operation in \cite{Pronobis2017,Peharz2019} need $\mathcal{O}(K^3)$ additions, $K^3$ exp-operations and $K$ $\log$-operations.
These implementations also store $3K$ values for $\node$, $\node'$, $\sumnode$, and additional $K^2$ values for the explicitly computed products.
While our implementation is cubic in the number of \emph{multiplications}, the previous implementations need a cubic number of \emph{exp-operations}. 
This partially explains the speedup of EiNets in our experiments in Section~\ref{sec:experiments}.
However, the main reasons for the observed speedup are i) an optimized implementation of the einsum operation, ii) avoiding the overhead of allocating product nodes, and iii) a higher degree of parallelizm, as discussed in the next section.

\subsection{The Einsum Layer}   \label{sec:einsum_layer}

Rather than computing single vectorized sums, we can do better by computing whole layers in parallel.
To this end, we first organize the PC in a layer-wise structure:
We traverse the PC top-down, and construct a topologically sorted list of layers of nodes, alternating between sums and products, and starting with the leaf nodes.
Nodes are only inserted in a layer if all their parents have already been inserted in some layer above, i.e.~all nodes in the $i^\text{th}$ layer depend only on nodes in layers with index strictly smaller than $i$.
Furthermore, note that since we can assume that each product has exactly one parent, see Section~\ref{sec:vectorizing}, a consecutive pair of product and sum layers will always be such that the product layer contains \emph{exactly} the inputs to the sum layer.
Pseudo-code for organizing PCs in topological layers is provided in the supplementary.
Our strategy for EiNets is to perform efficient parallel computations for i) the whole leaf layer, and ii) the whole sum layer in each pair of consecutive product and sum layers.
In both cases, the result is a matrix of log-densities, with as many rows as there are leafs (resp.~sums) in the layer, and $K$ columns.

\begin{figure}[t]
\centering
\includegraphics[width=0.3\textwidth]{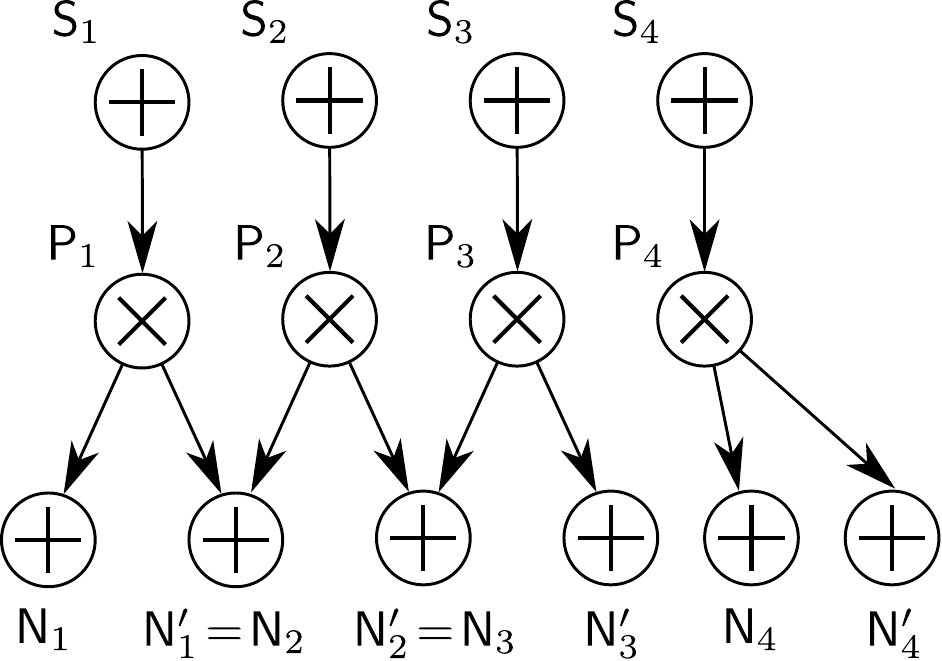}%
\caption{Example of an einsum layer, parallelizing the basic einsum operation.}
\label{fig:basic_einsum_layer}
\end{figure}

Computing the input layer is discussed in Section~\ref{sec:ef_input_layer}.
For now, consider a layer of $L$ sum nodes, as illustrated in Fig.~\ref{fig:basic_einsum_layer}.
Here, we have assumed that each sum node has only a single child.
We discuss this simpler case first, and handle sum nodes with multiple children below.
In order to compute the $L$ sums, we collect all the vectors of the ``left'' product children in a $L \times K$ matrix $\bm{\node}$ and similarly all the ``right'' product children in $L \times K$ matrix $\bm{\node}'$, where ``left'' and ``right'' are arbitrary but fixed assignments. 
We further extend the 3-dimensional weight-tensor $\mathbf{W}$ from \eqref{eq:basic_einsum} to a 4-dimensional $L \times K \times K \times K$ tensor, where the slice $\mathbf{W}_{l\colon\!\!\colon\!\!\colon}$ contains the weights for the $l^\text{th}$ vectorized sum node.
The result of all $L$ sums---i.e.~in total $L \times K$ sum operations---can be performed with a single einsum operation:
\begin{equation}    \label{eq:layer_einsum}
\bm{\sumnode}_{lk} = \mathbf{W}_{lkij} \bm{\node}_{li} \bm{\node'}_{lj} , 
\end{equation}
Note the similarity to \eqref{eq:basic_einsum}, and that we parallelized over the whole sum layer by simply introducing an additional index $l$.
Consequently, it is straight-forward to extend the log-einsum-exp trick (see Section \eqref{sec:basic_einsum}) to \eqref{eq:layer_einsum}.
Constructing the two matrices $\bm{\node}$ and $\bm{\node}'$ requires some book-keeping and introduces some computational overhead stemming from extracting and concatenating slices from the log-probability tensors below.
This overhead is essentially the symptom of the sparse and cluttered layout of PCs.
The main computational work, however, is then performed by a highly parallel einsum operation \eqref{eq:layer_einsum}.

Eq.~\eqref{eq:layer_einsum} computes whole sum-product-layers, when the sum nodes are restricted to single children.
In order to compute general layers, we express any layer containing sums with multiple children as 2 consecutive sum layers.
The first layer consists of sum nodes with single children, one for each product child of the original sum layer. 
The second layer takes element-wise mixtures of the sum nodes in the first layer, over sum nodes which correspond to children of the original layer. 
Note that this structure is simply an over-parameterization \cite{Trapp2019b} of the original sum nodes, decomposing them into chains of two sum nodes.
The first layer can now be efficiently computed with \eqref{eq:layer_einsum}.
The second layer, which we denote as \emph{mixing layer}, can also be computed in parallel with an einsum operation.
For details, see the supplementary paper.

\subsection{Exponential Families as Input Layer}   \label{sec:ef_input_layer}

The leaves of EiNets compute log-densities of an exponential family (EF), which has the form
$
\log \leaf = \log h(\x) + T(\x)^T \theta - A(\theta),
$
where $\theta$ are the natural parameters, $h$ is the so-called base measure, $T$ the sufficient statistic and $A$ is the log-normalizer.
Many parametric distributions can be expressed as EFs, e.g.~Gaussian, Binomial, Categorical, Beta, etc.
In order to facilitate learning using EM, we keep the parameters in their \emph{expectation} form $\phi$ \cite{Sato1999}.
The natural parameters $\theta$ and expectation parameters $\phi$ are one-to-one, and connected via $\phi = \partial A(\theta) / \partial \theta$ and $\theta = \partial H(\phi) / \partial \phi$, where $H$ denotes entropy.
This dual parameterization allows us to implement EM on the abstract level of EFs, while particular instances of EFs are easily implemented by providing $h$, $T$, $A$, and the conversion $\theta(\phi)$.

In order to compute all leaves in parallel, we first compute a $D \times K \times R$ tensor $\mathbf{E}$ of EF log-probabilities, where $D$ is the number of RVs and $R$ is the number of so-called \emph{replica}.
Each entry $E_{d,k,r}$ is a log-density of an EF over $X_d$.
The number of replica must be large enough in order to ensure that each leaf has a set of $K$ private EF distributions, i.e.~each leaf has an index $r$ to a particular replica, such that leaves with the same $r$ have disjoint scope.
The number of required replica $R$ and an assignment of leaves to replica can  be inferred from the PC's structure.
The EF tensor $\mathbf{E}$ is parameterized with a $D \times K \times R \times |T|$ tensor, where $|T|$ is the dimensionality of the sufficient statistic $T$.
Note that $\mathbf{E}$ can be computed in parallel with a handful of parallel operations, e.g.~inner product $T(\x)^T \theta$, evaluating $A(\theta)$, etc.
$\mathbf{E}$ is then used to compute the leaves, which, in this paper, are simply factorizations over the log-densities in $\mathbf{E}$.

\subsection{Expectation-Maximization (EM)} \label{sec:em}

A natural way to learn PCs is the EM algorithm, wich is known to rapidly increase the likelihood, especially in early iterations \cite{Salakhutdinov2003}.
EM for PCs was derived in \cite{Peharz2017}, leading to the following update rules for sum-weights and leaves:
\begin{align}
\label{eq:em_statistics}
& n_{\sumnode,\node}(\x) = 
\frac{1}{\PC(\x)} \frac{\partial \PC}{\partial \sumnode} \node(\x),
~~~
p_{\leaf}(\x) = \frac{1}{\PC(\x)} \frac{\partial \PC}{\partial \leaf} \leaf(\x),
\\
\label{eq:em_update}
& w_{\sumnode,\node} \leftarrow \frac{w_{\sumnode,\node} \, \sum_\x n_{\sumnode,\node}(\x)}{\sum_{\x,\node \in \ch(\sumnode)} n_{\sumnode,\node}(\x)},
~~~
\phi_{\leaf} \leftarrow \frac{\sum_\x p_{\leaf(\x)} T(\x)}{\sum_\x p_{\leaf(\x)}},
\end{align}
where the sums in \eqref{eq:em_update} range over all training examples $\x$.
In \cite{Peharz2017} and \cite{Peharz2019}, the derivatives $\frac{\partial \PC}{\partial \sumnode}$ and $ \frac{\partial \PC}{\partial \leaf}$ required for the expected statistics $n_{\sumnode,\node}$ and $p_{\leaf}$ were computed with an \emph{explicitly implemented backwards-pass}, performed in the log-domain for robustness.
Here we show that this implementation overhead can be avoided by leveraging automatic differentiation.
Recall that EiNets represent all probability values in the log-domain, thus the PC output is actually $\log \PC(\x)$, rather than $\PC(\x)$.
Calling auto-diff on $\log \PC(\x)$ yields the following derivative for each sum-weight $w_{\sumnode,\node}$ (omitting argument $\x$):
$
\frac{\partial \log \PC}{w_{\sumnode,\node}} 
 = \frac{1}{\PC} \frac{\partial \PC}{\partial\sumnode} \frac{\partial \sumnode}{\partial w_{\sumnode,\node}}  
= \frac{1}{\PC} \frac{\partial \PC}{\partial\sumnode} \node,
$
which is exactly $n_{\sumnode,\node}$ in \eqref{eq:em_statistics}, i.e.~auto-diff readily provides the required expected statistics for sum nodes.
In many frameworks, auto-diff readily accumulate the gradient by default, as required in \eqref{eq:em_update}, leading to an embarrassingly simple implementation for the E-step.
For the M-step, we simply multiply the result of the accumulator with the current weights, and renormalize.
Furthermore, recall that each single-dimensional leaf is implemented as log-density of an EF.
Taking the gradient yields:
$
\frac{\partial \log \PC}{\partial \log L} 
= \frac{1}{\PC} \frac{\partial \PC}{\partial \log L}  
= \frac{1}{\PC} \frac{\partial \PC}{\partial L} L,
$
which is $p_\leaf$ in \eqref{eq:em_statistics}.
Thus, auto-diff also implements most of the EM update for the leaves.
We simply need to accumulate both $p_\leaf$ and $p_\leaf \, T$ over the whole dataset, and use \eqref{eq:em_update} to update the expectation parameters $\phi$.
Note, that both sum nodes and leafs can be updated using the same calls to auto-diff.

The classical EM algorithm uses a whole pass over the training set for a single update, which is computationally wasteful due to redundancies in the training data \cite{Bottou1998}.
Similar to stochastic gradient descent (SGD), it is possible to define a stochastic version for EM \cite{Sato1999}.
To this end, one replaces the sums over the entire dataset in \eqref{eq:em_update} with sums over \emph{mini-batches}, yielding $\w^{mini}_{\sumnode,\node}$ and $\phi^{mini}_{\leaf}$.
The full EM update is then replaced with gliding averages
\begin{align}
\label{eq:stochatic_em_w}
\w_{\sumnode,\node} & \leftarrow (1-\lambda) \, \w_{\sumnode,\node} + \lambda \, \w^{mini}_{\sumnode,\node}   \\
\label{eq:stochatic_em_leaf}
\phi_{\leaf}  & \leftarrow (1-\lambda) \, \phi_{\leaf} + \lambda \, \phi^{mini}_{\leaf},
\end{align}
where $\lambda \in [0,1]$ is a step-size parameter.
This \emph{stochastic} version of EM introduces two hyper-parameters, step-size $\lambda$ and the batch-size, which need to be set appropriately.
Furthermore, unlike full-batch EM, stochastic EM does not guarantee that the training likelihood increases in each iteration.
However, stochastic EM updates the parameters after each mini-batch and typically leads to faster learning.

\citet{Sato1999} shows an interesting connection between stochastic EM and natural gradients \cite{Amari1998}.
In particular, for any EF model $P(\X, \Z)$, where $\X$ are observed and $\Z$ latent variables, performing \eqref{eq:stochatic_em_w} and \eqref{eq:stochatic_em_leaf} is equivalent to SGD, where the gradient is pre-multiplied by the inverse Fisher information matrix of the \emph{complete} model $P(\X, \Z)$.
The difference to standard natural gradient is, that the natural gradient is defined via the inverse Fisher of the \emph{marginal} $P(\X)$ \cite{Amari1998}.
Sato's analysis also applies to EiNets, since smooth and decomposable PCs \emph{are} EFs of the form $P(\X, \Z)$ \cite{Peharz2017}, where the latent variables are associated with sum nodes.
Thus, \eqref{eq:stochatic_em_w} and \eqref{eq:stochatic_em_leaf} are in fact performing SGD with (a variant of) natural gradient.

\section{Experiments}   \label{sec:experiments}

\begin{table}
\scriptsize
\centering
\caption{
Sanity check that EiNets reproduce the test log-likelihood of RAT-SPNs \cite{Peharz2019} when trained on 20 binary datasets, and using the same structures.
On most datasets, the results between RAT-SPNs and EiNets are not significantly different (results on boldface, using a one sided t-test, $p=0.05$). 
}
\setlength{\tabcolsep}{1.5pt}
\begin{tabular}{l rr}
\toprule
dataset       &                         RAT-SPN   &                          EiNet      \\
\midrule
nltcs         &                        \textbf{\underline{-6.011}}  &                \textbf{-6.015}      \\
msnbc         &                        \textbf{\underline{-6.039}}  &                         -6.119      \\
kdd-2k        &                        \textbf{\underline{-2.128}}  &                \textbf{-2.183}      \\
plants        &                        \textbf{\underline{-13.439}}  &               \textbf{-13.676}      \\
jester        &                        \textbf{-52.970}  &   \textbf{\underline{-52.563}}      \\
audio         &                        \textbf{-39.958}  &   \textbf{\underline{-39.879}}      \\
netflix       &                        \textbf{-56.850}  &   \textbf{\underline{-56.544}}      \\
accidents     &                       \textbf{\underline{-35.487}}  &               \textbf{-35.594}      \\
retail        &                         \textbf{\underline{-10.911}}  &               \textbf{-10.916}      \\
pumsb-star    &                         \textbf{-32.530}  &   \textbf{\underline{-31.954}}      \\
dna           &                         -97.232  &   \textbf{\underline{-96.086}}      \\
kosarek       &                         \textbf{\underline{-10.888}}  &               \textbf{-11.029}      \\
msweb         &                         \textbf{-10.116}  &   \textbf{\underline{-10.026}}      \\
book          &                         \textbf{\underline{-34.684}}  &               \textbf{-34.739}      \\
each-movie    &                         \textbf{-53.632}  &   \textbf{\underline{-51.705}}      \\
web-kb        &                         \textbf{-157.530}  &  \textbf{\underline{-157.282}}      \\
reuters-52    &                         \textbf{\underline{-87.367}}  &               \textbf{-87.368}      \\
20ng          &                         \textbf{\underline{-152.062}}  &              \textbf{-153.938}      \\
bbc           &                         \textbf{-252.138}  &  \textbf{\underline{-248.332}}      \\
ad            &                         -48.472  &   \textbf{\underline{-26.273}}      \\
\bottomrule
\end{tabular}
\label{tab:density-estimation}
\end{table}

\begin{figure*}[ht]
\centering
\includegraphics[scale=0.2]{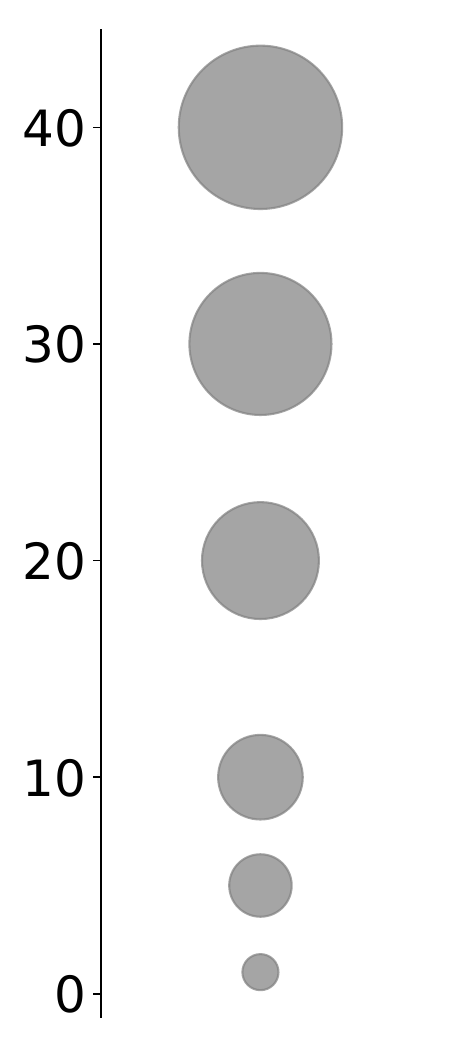}%
~
\includegraphics[scale=0.2]{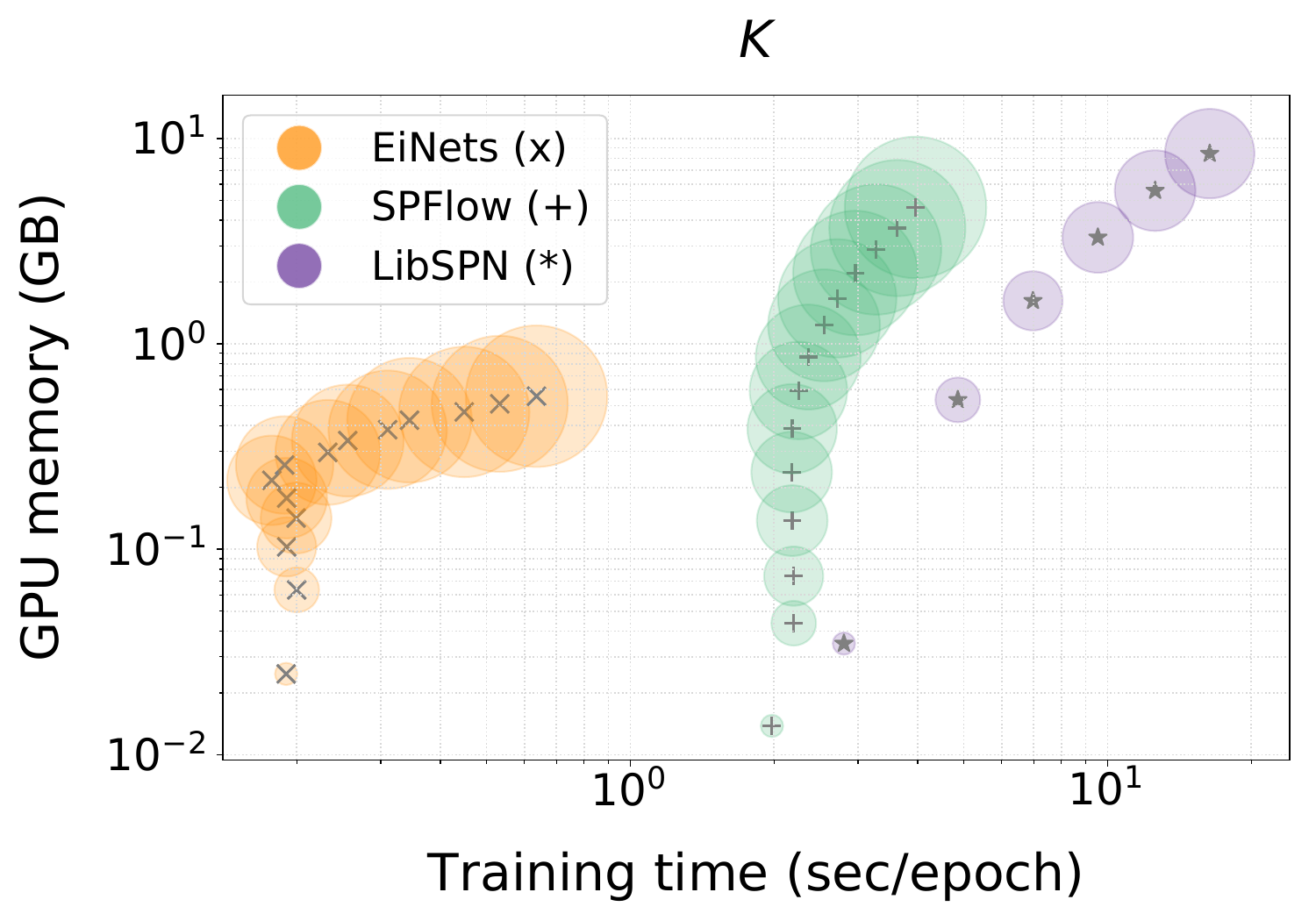}%
\includegraphics[scale=0.2]{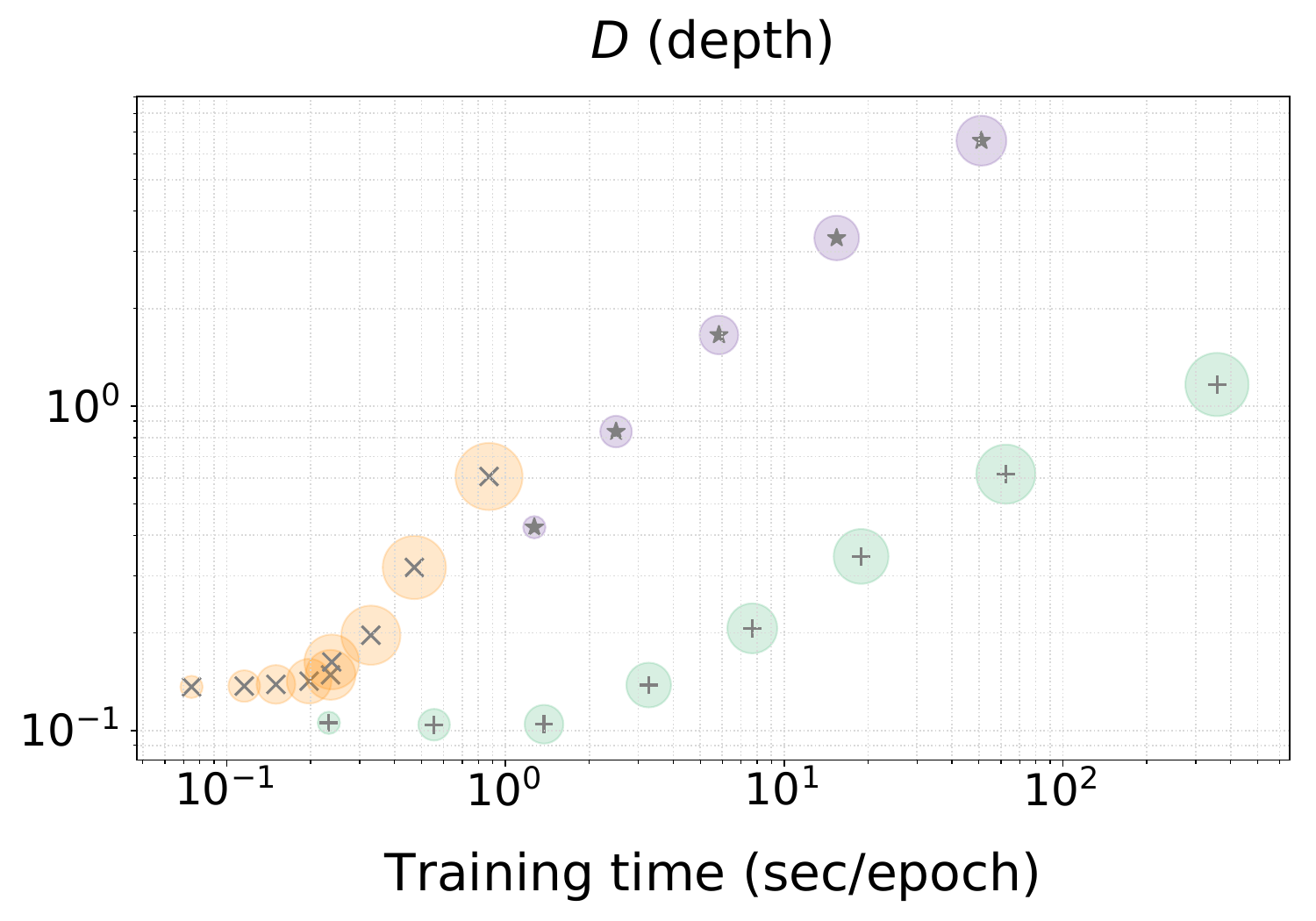}%
\includegraphics[scale=0.2]{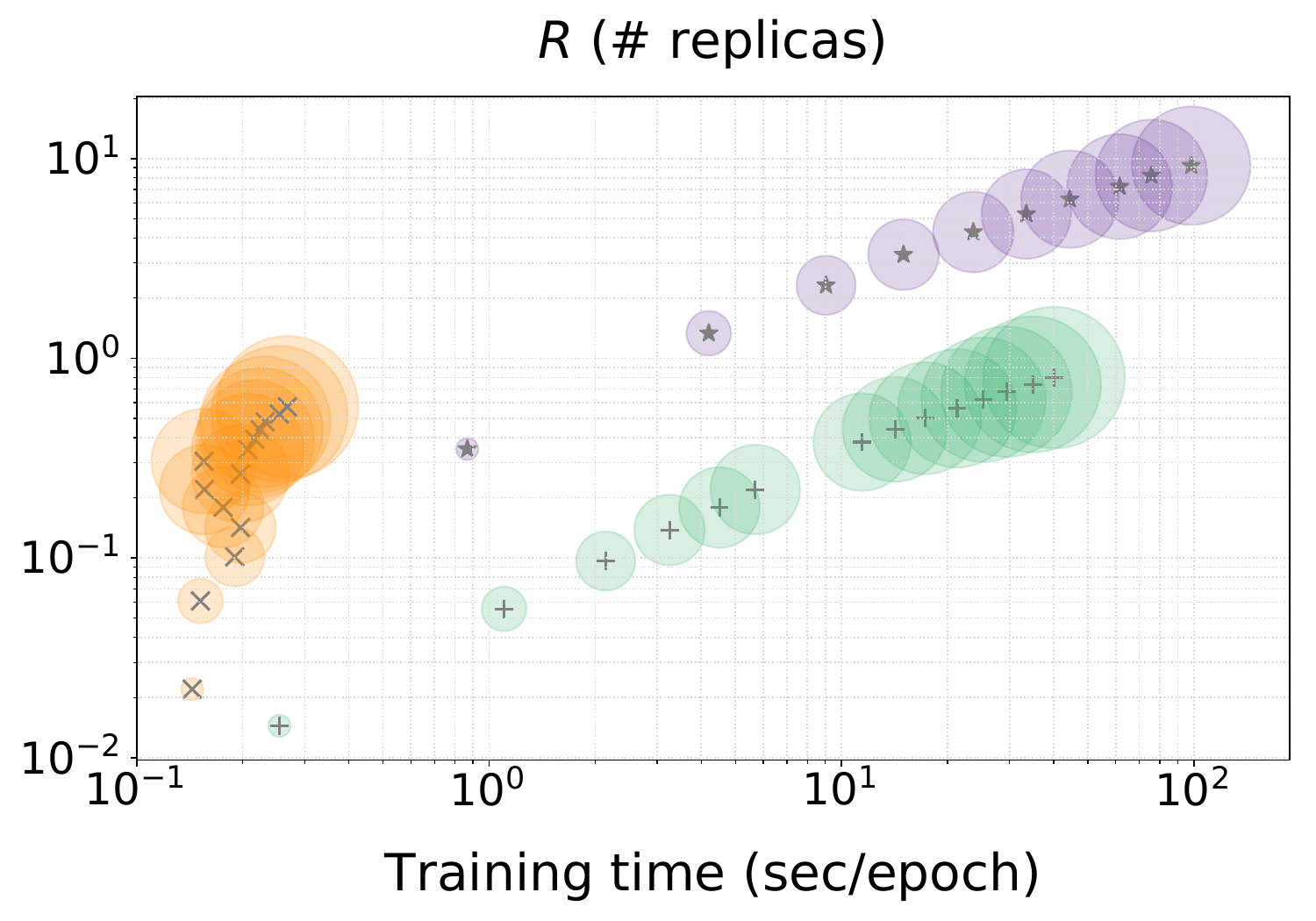}%
\caption{Illustration of training time and peak memory consumption of EiNets, SPFlow and LibSPN when training randomized binary PC trees, and varying hyper-parameters $K$ (number of densities per sum/leaf), depth $D$, and number of replica $R$, respectively.
The blob size directly corresponds to the respective hyper-parameter under change.
The total number of parameters ranged within 
$10k-9.4M$ (for varying $K$),
$100k-5.2M$ (for varying $D$), and
$24k-973k$ (for varying $R$).
For LibSPN, some settings exhausted GPU memory and are therefore missing.
} 
\label{fig:efficiency_comparison}
\end{figure*}

We implemented EiNets fully in PyTorch \cite{Paszke2019}.
The core implementation is provided in the supplementary, and we will release the code in the near future.

\subsection{Efficiency Comparison}   \label{sec:efficiency}

In order to demonstrate the efficiency of EiNets we compare with the two most prominent ``deep-learning'' implementations of PCs, namely LibSPN \cite{Pronobis2017} and SPFlow \cite{Molina2019}.
LibSPN is natively based on Tensorflow \cite{Abadi2015}, while SPFlow supports multiple backends.
For our experiment, we used SPFlow with Tensorflow backend.
We used randomized binary trees (RAT-SPNs) \cite{Peharz2019} as a common benchmark:
These PC structures are governed by two structural parameters, the \emph{split-depth} $D$ and \emph{number of replica} $R$:
Starting from the root sum node, they split the whole scope $\X$ into two randomized balanced parts, recursively until depth $D$, yielding a binary tree shape with $2^D$ leaves.
This construction is repeated $R$ times, yielding $R$ random binary trees, which are mixed at the root.
As a sanity check, we first reproduced the density estimation results on 20 binary datasets in \cite{Peharz2019}, see Table~\ref{tab:density-estimation}. 
The difference in test log-likelihood is for most of the datasets not statistically significant, using a one-sided t-test with $p=0.05$.  
EiNets even outperformed RAT-SPNs on two datasets, once with a large margin, while RAT-SPNs outperformed EiNets on one datset, with a small margin.

We aim to compare EiNets, LibSPN and SPFlow in terms of i) training time, ii) memory consumption, and iii) inference time.
To this end, we trained and tested them on synthetic data (Gaussian noise) with $N=2000$ samples and $D=512$ dimensions.
As leaves we used single-dimensional Gaussians.
We varied each structural hyper-parameter in the following ranges: depth $D \in \{1,\dots,9\}$, replica $R \in \{1,\dots,40\}$, and vector length of sums/leaves $K\in\{1,\dots,40\}$.
When varying one hyper-parameter, we left the others to default values $D=4$, $R=10$, and $K=10$.
We ran this set of experiments on a GeForce RTX $2080$ Ti.

In Figure~\ref{fig:efficiency_comparison} we see a comparison of the three implementations in terms of train time and GPU-memory consumption, where the circle radii represent the magnitude of the varied hyper-parameter.  
We see that EiNets tend to be \textbf{one or two orders of magnitude} faster (note the log-scale) than the competitors, especially for large models.  
Also in terms of memory consumption EiNets scale gracefully. 
In particular, for large $K$, memory consumption is an order of magnitude lower than for LibSPN or SPFLow.
This can be easily explained by the fact that our einsum operations do not generate product nodes explicitly in memory, while the other frameworks do.
Moreover, EiNets also perform superior in terms of inference time.
In particular for large models, they run again one or two orders of magnitude faster than the other implementations.
For space reasons, these results are deferred to the supplementary.

\begin{figure*}
\centering
\begin{subfigure}{.24\textwidth}
\includegraphics[width=0.99\textwidth]{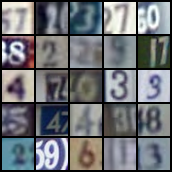}
\caption{Real SVHN images.}
\end{subfigure}
~
\begin{subfigure}{.24\textwidth}
\includegraphics[width=0.99\textwidth]{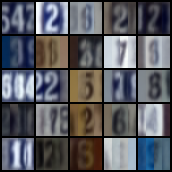}
\caption{EiNet SVHN samples.}
\end{subfigure}
~
\begin{subfigure}{.477\textwidth}
\includegraphics[width=0.99\textwidth]{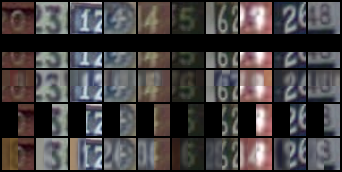}
\caption{Real images (top), covered images, and EiNet reconstructions}
\end{subfigure}
\\
\begin{subfigure}{.24\textwidth}
\includegraphics[width=0.99\textwidth]{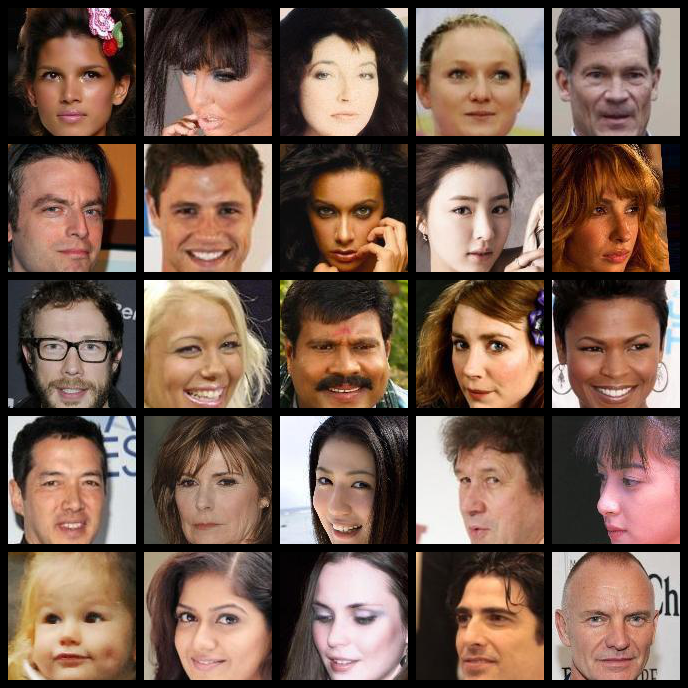}
\caption{Real Celeba samples.}
\end{subfigure}
~
\begin{subfigure}{.24\textwidth}
\includegraphics[width=0.99\textwidth]{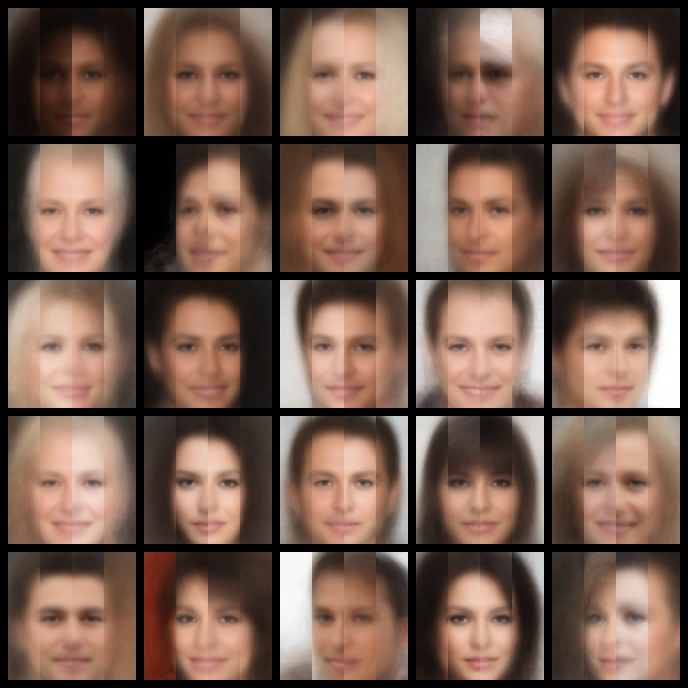}
\caption{EiNet Celeba samples.}
\end{subfigure}
~
\begin{subfigure}{.477\textwidth}
\includegraphics[width=0.99\textwidth]{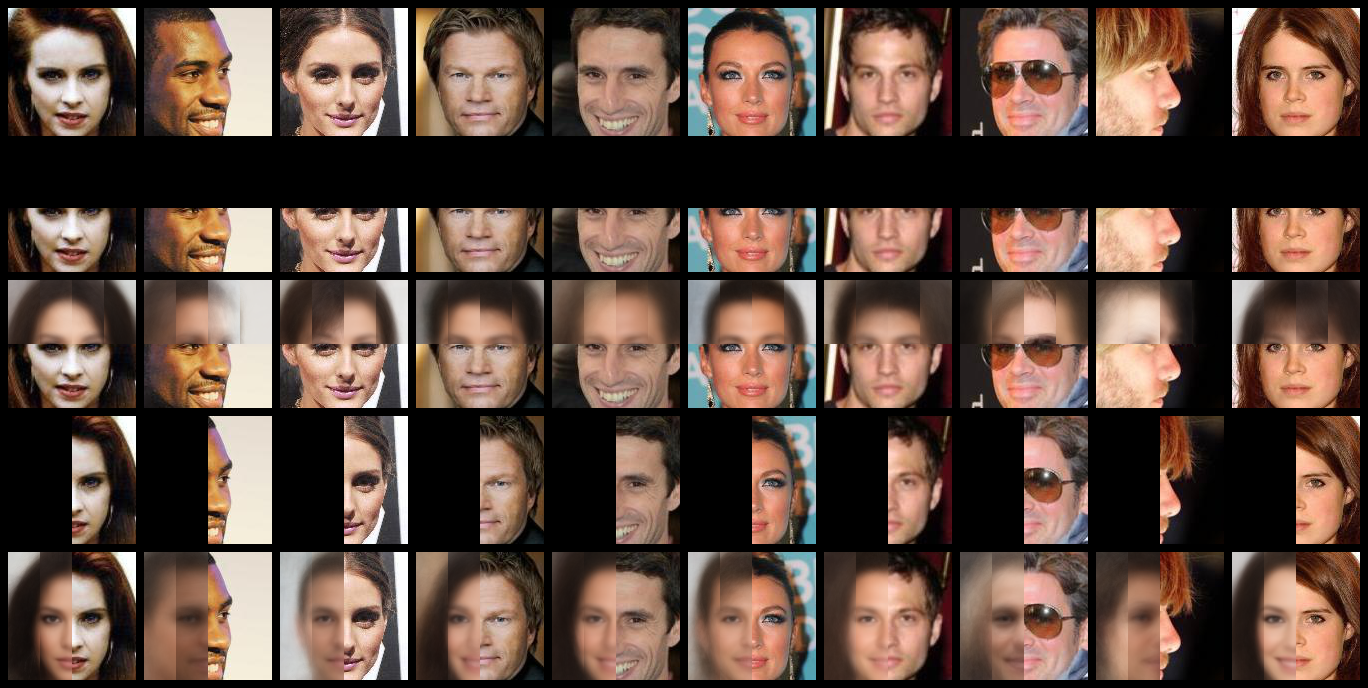}
\caption{Real images (top), covered images, and EiNet reconstructions}
\end{subfigure}
\caption{Qualitative results of EiNets trained on RGB data, namely SVHN (top, image dimensions $32 \times 32$) and Celeba (bottom, image dimensions $128 \times 128$).
In all samples, the means of the Gaussian leaves were used---see Section~\ref{sec:addendum} for more information.
}
\label{fig:celeba}
\vspace{-0.5\baselineskip}
\end{figure*}

\subsection{EiNets as Generative Image Models}

While PCs are an actively researched topic, their scalability issues have restricted them so far to rather small datasets like MNIST \cite{Lecun1998}.
In this paper, we use PCs as generative model for street-view house numbers (SVHN) \cite{Netzer2011}, containing $32 \times 32$ RGB images of digits, and center-cropped CelebA \cite{Liu2015}, containing $128 \times 128$ RGB face images.
For SVHN, we used the first $50k$ train images and concatenated it with the extra set, yielding a train set $581k$ images.
We reserved the rest of the core train set, $23k$ images, as validation set. 
The test set comprises $26k$ images. 
For Celeba, we used the standard train, validation and test splits, containing $183k$, $10k$, and $10k$, respectively.
The data was normalized before training (division by $255$), but otherwise no preprocessing was applied. 
To the best or our knowledge, PCs have not been successfully trained on datasets of this size before.

We trained EiNets on the image-tailored structure proposed in \cite{Poon2011}, to which we refer as PD structure. 
The PD structure recursively decomposes the image into sub-rectangles using axis-aligned splits, displaced by a certain step-size $\Delta$.
Here, $\Delta$ serves as a structural hyperparameter, which governs the number of sum nodes according to $\mathcal{O}(\frac{1}{\Delta^3})$ \cite{Peharz2015a}.
The recursive splitting process stops, when a rectangle cannot be split by any value in $\Delta$.
We first clustered both datasets into $100$ clusters using the \emph{sklearn} implementation of k-means, and learned an EiNet on each of these clusters.
We then used these $100$ EiNets as mixture components of a mixture model, using the cluster proportions as mixture coefficients.
Note that i) a mixture of PCs yields again a PCs, and ii) this step is essentially the first step of \emph{LearnSPN} \cite{Gens2013}, one of the most prominent PC structure learners.
For the PD structure of each EiNet component, we used a step-size $\Delta=8$ for SVHN and $\Delta=32$ for CelebaA, i.e.~we applied $4$ axis-aligned splits.
We only applied vertical splits, in order to reduce artifacts stemming from the PD structure.
As leaves, we used factorized Gaussians, i.e.~Gaussians with diagonal covariance mattrix, which were further factorized over the RGB channels.
The vector length for sums and leaves was set to $K=40$.
After each EM update, we projected the Gaussian variances to the interval $[10^{-6}, 10^{-2}]$, corresponding to a maximal standard deviation of $0.1$.
Each component was trained for $25$ epochs, using a batch size of $500$ and EM stepsize of $0.5$.
In total, training lasted $5$ hours for SVHN and $3$ hours for CelebA, on an NVidia P100.

Original samples and samples from the EiNet mixture are shown in Figure~\ref{fig:celeba}, (a,b) for SVHN and (d,e) for CelebA, respectively.
The SVHN samples are rather compelling, and some samples could be mistaken for real data samples.
The CelebA samples are somewhat over-smoothed, but captured the main quality of faces well.
In both cases, some ``stripy'' artifacts, typical for the PD architecture \cite{Poon2011}, can be seen.
Although the image quality is not comparable to e.g.~GAN models \cite{Goodfellow2014}, we want to stress that EiNets permit tractable inference, while GANs are restricted to sampling.
In particular, tractable inference can be immediately used for image inpainting, as demonstrated in Figure~\ref{fig:celeba}, (c) for SVHN and (f) for CelebA.
Here, we marked parts of test samples as missing (top or left image half) and reconstructed it by drawing a sample from the conditional distribution, conditioned on the visible image part. 
We see that the reconstructions are plausible, in particular for SVHN.

\section{Conclusion}    \label{sec:conclusions}

Probabilistic models form a spectrum of machine learning techniques.
Most of the research is focused on representing and learning flexible and expressive models, but ignore the down-stream impact on the set of inference tasks which can be provably solved within the model.
The philosophy of tractable modeling also aims to push the expressivity boundaries, but under the constraint to maintain a defined set of exact inference routines.
Probabilistic circuits are certainly a central and prominent approach for this philosophy.
In this paper, we addressed a major obstacle for PCs, namely their scalability in comparison to unconstrained models.
Our improvements of training speed and memory-use reduction, both in the orders of one or two orders of magnitude, are compelling, and we hope that our results stimulates further developments in the area of tractable models.

\appendix

\section{Organizing EiNets in Topological Layers}

In this section, we elaborate on the layer-wise organization of EiNets, as discussed in Section 3.3 in the main paper.
The layer-wise organization is obtained by Algorithm~\ref{alg:layered_pc}, which takes the computational graph $\graph$ of some EiNet as input, and which performs a type of breadth-first search over $\graph$.

We first divide the node set $V$ into the node sets $\bm{\leaf}$, $\bm{\sumnode}$, and $\bm{\prodnode}$, containing all leaves, all sums and all products, respectively.
We initialize an empty set $M$, which will contain all ``visited'' nodes during the execution of the algorithm.
We also initialize an empty list $layers$ which will be a topologically ordered list of pure node-sets (i.e.~containing exclusively nodes form $\bm{\leaf}$, $\bm{\sumnode}$, or $\bm{\prodnode}$), and which will contain the result of the Algorithm.

The while-loop in Algorithm~\ref{alg:layered_pc} is executed until all sum and product nodes have been visited and stored in some layer.
In line 6, we construct the set $l_\sumnode$ of sum nodes $\sumnode$ which have not been visited yet $\sumnode \notin M$, but whose parents have all been visited.
Note that in the first iteration of the while loop $l_\sumnode$ will simply contain the root of the EiNet.
The set $l_\sumnode$ is then inserted in the front of the list $layers$ in line 7, and all nodes in $l_\sumnode$ are marked as visited in line 8.
A similar procedure, but for product nodes, is performed in lines 9--11 yielding node set $l_\prodnode$.
Note that since we assume that each product node has only one parent (cf.~Section 3.1 in the main paper), $l_\prodnode$ will be \emph{exactly} the set of children of sum nodes contained in $l_\sumnode$ (constructed in line 6 within the same while-iteration).
Thus, if additionally all sum nodes have exactly one child, $l_\prodnode$ and $l_\sumnode$ will form a consecutive pair of product and sum nodes, as illustrated in Fig.~2 in the main paper, which can be compactly computed using a single call to an einsum operation.
If sum nodes have multiple children, we decompose them into chains of $2$ consecutive sum nodes, as discussed in the next section.

In line 13 of Algorithm~\ref{alg:layered_pc}, all leaf nodes are inserted as the bottom layer.
It is easy to check that the returned list $layers$ will be topologically sorted, meaning that the nodes in the $i^\text{th}$ layer will only have inputs from layers with index strictly smaller than $i$.

\begin{algorithm}[t]
   \caption{Topological Layers}
   \label{alg:layered_pc}
\begin{algorithmic}[1]
   \STATE {\bfseries Input:} PC graph $\graph = (V,E)$
   \STATE Let $\bm{\leaf}$, $\bm{\sumnode}$, $\bm{\prodnode}$ 
   be the set of all leaves, sum nodes, product nodes in $V$, respectively
   \STATE $M \leftarrow \{\}$  
   \STATE $layers = [ \, ]$
   \WHILE{$M \not= \bm{\sumnode} \cup \bm{\prodnode}$}
   \STATE $l_\sumnode = \{\sumnode \in \bm{\sumnode} \,|\, \sumnode \notin M \land \forall \prodnode \in \pa(\sumnode)\colon \prodnode \in M\}$
   \STATE $layers \leftarrow concatenate([l_\sumnode], layers)$
   \STATE $M \leftarrow M \cup l_\sumnode$
   \STATE $l_\prodnode = \{\prodnode \in \bm{\prodnode} \,|\, \prodnode \notin M \land \forall \sumnode \in \pa(\prodnode)\colon \sumnode \in M\}$
   \STATE $layers \leftarrow concatenate([l_\prodnode], layers)$
   \STATE $M \leftarrow M \cup l_\prodnode$
   \ENDWHILE
   \STATE $layers \leftarrow concatenate([\bm{\leaf}], layers)$ 
   \STATE {\bfseries Return} $layers$
\end{algorithmic}
\end{algorithm}

\section{The Mixing Layer}   \label{sec:mixing_layer}

\begin{figure}
\centering
\includegraphics[width=0.48\textwidth]{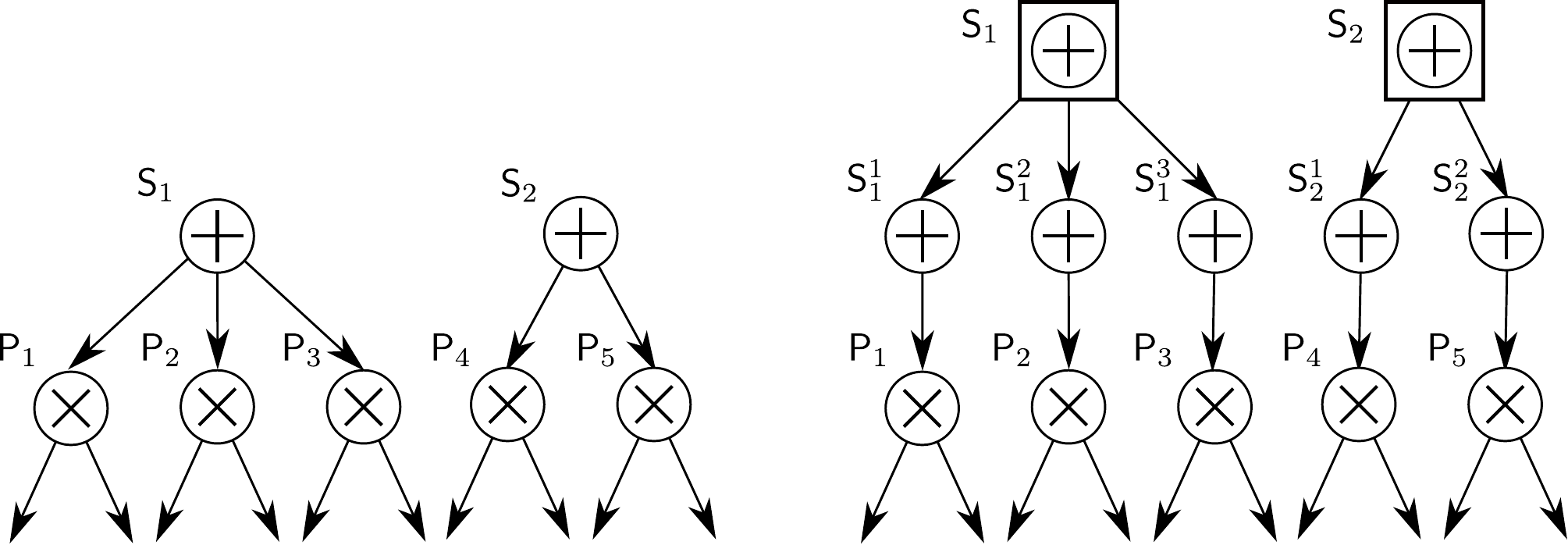}%
\caption{Decomposing a layer of sum nodes with multiple children (left) into two consecutive sum layers (right).
The first sum layer computes a standard einsum layer, discussed in Section~3.3 in the main paper.
The second layer, the so-called mixing layer, takes element-wise mixtures (depicted with sums in boxes).
}
\label{fig:mixing_layer}
\end{figure}
\begin{figure*}[t!]
\centering
\includegraphics[scale=0.2]{radius.pdf}%
~
\includegraphics[scale=0.2]{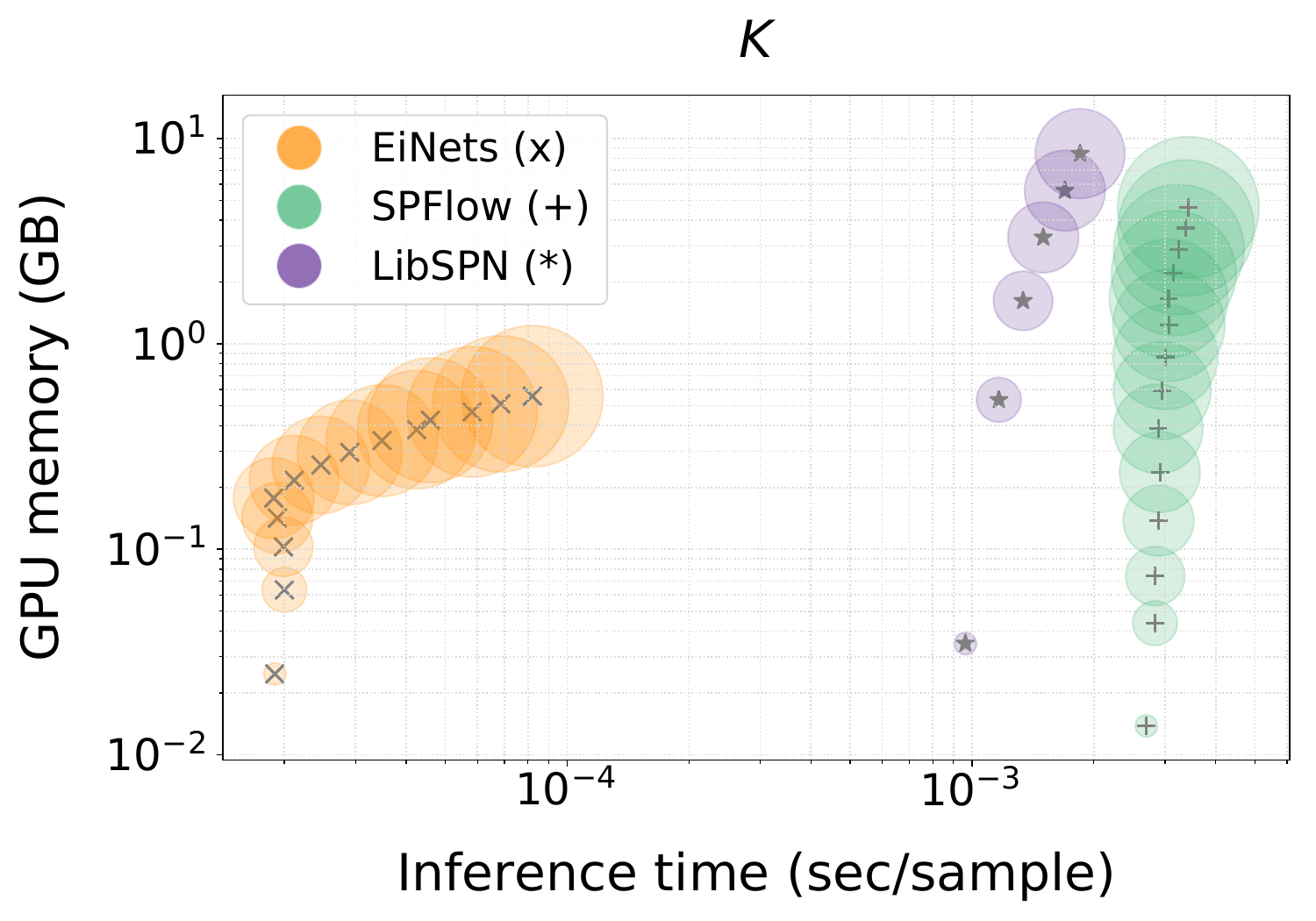}%
\includegraphics[scale=0.2]{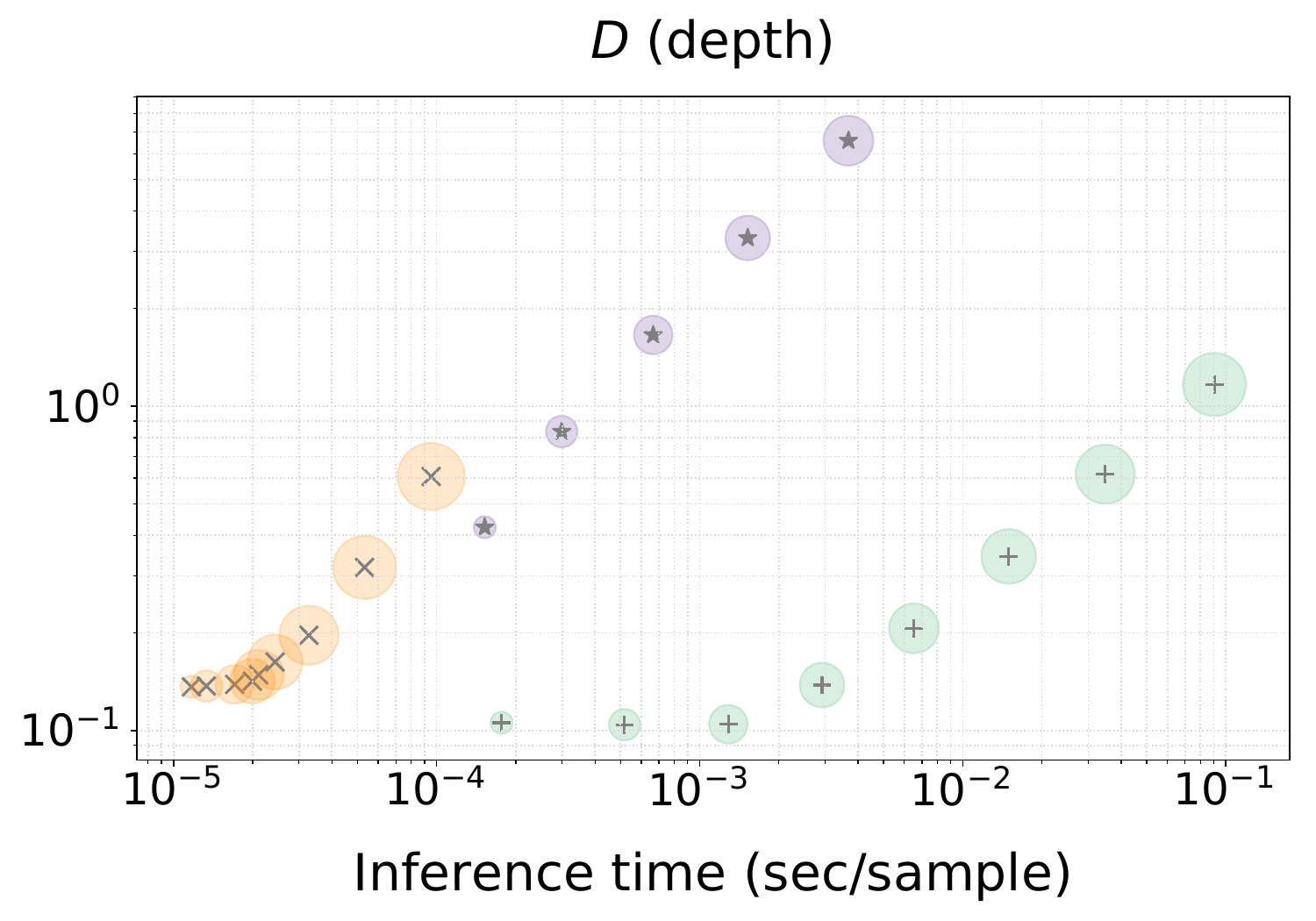}%
\includegraphics[scale=0.2]{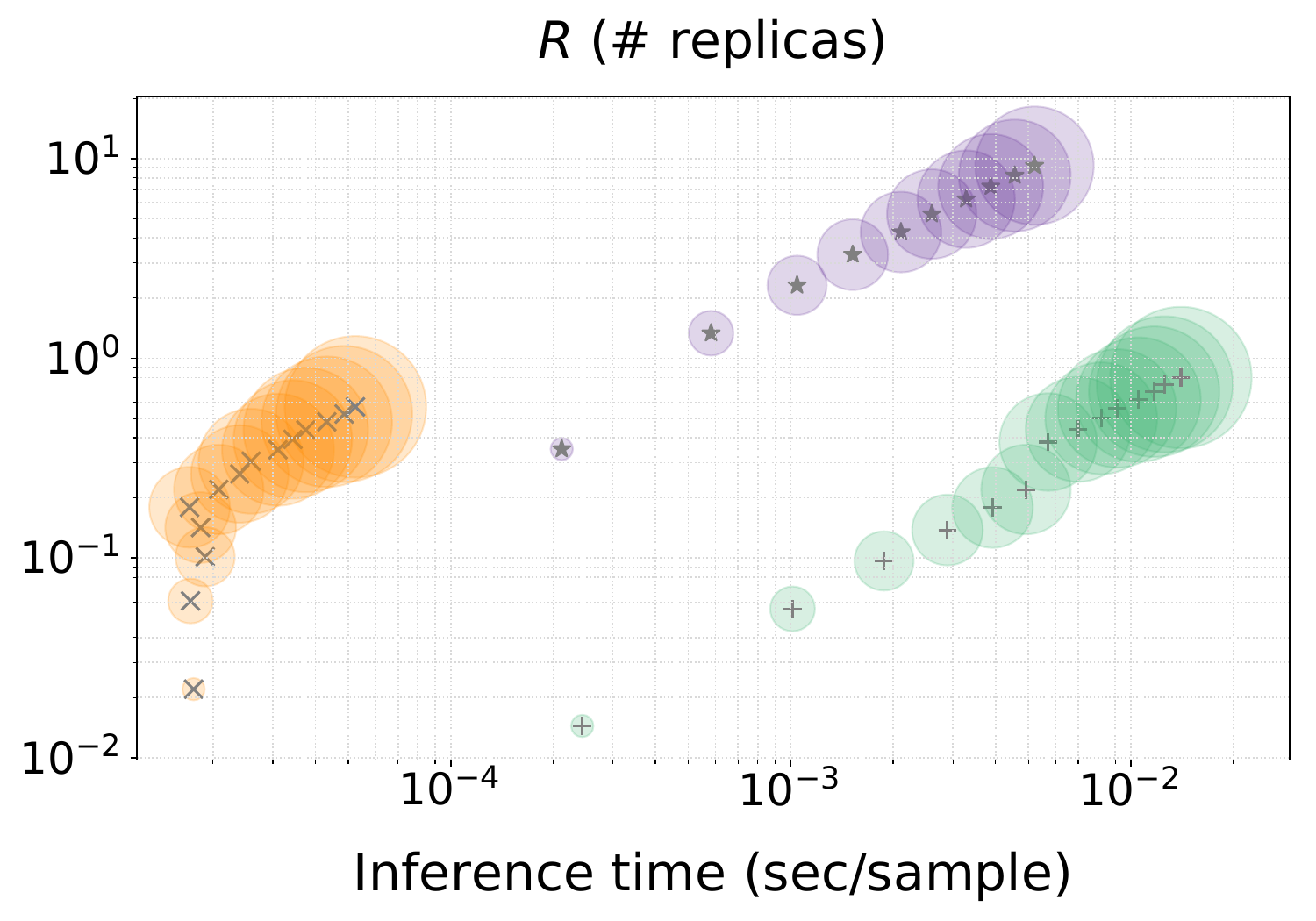}%
\caption{Illustration of inference time and peak memory consumption of EiNets, SPFlow and LibSPN on randomized binary PC trees, and varying hyper-parameters $K$ (number of densities per sum/leaf), depth $D$, and number of replica $R$, respectively.
The blob size directly corresponds to the respective hyper-parameter under change.
The unchanged hyper-parameters were fixed to their respective default values:  $D=4$, $R=10$, and $K=10$.
The total number of parameters ranged within 
$10k-9.4M$ (for varying $K$),
$100k-5.2M$ (for varying $D$), and
$24k-973k$ (for varying $R$).
For LibSPN, some settings exhausted GPU memory and are therefore missing.
} 
\label{fig:efficiency_comparison_supp}
\end{figure*}

The einsum layer computes a large number of sum-product operations with a single call to an einsum operation, but requires that each sum node has exactly one child.
In order to compute sum nodes with multiple product children, we express any sum node with multiple children as a cascade of 2 vectorized sum operations.
In particular, for a sum node $\sumnode$ with $C$ children, we introduce $C$ new sum nodes $\sumnode^1, \dots, \sumnode^C$, each having one of the children as its single child.
This forms a layer of sum nodes with single children, which can be computed with a single einsum operation depicted in Eq.~(5) in the main paper.
Subsequently, the results of $\sumnode^c$, $c\in\{1,\dots,C\}$, get mixed in an element-wise manner, 
i.e.~$\sumnode_k = \sum_{c=1}^C w^c \sumnode_k^c$.
We call the $\sumnode^c$ simple sums and $\sumnode$ aggregated sum.

This structure is simply an over-parameterization \cite{Trapp2019b} of the original sum nodes, and represents the same linear functions.  
This principle is illustrated in Fig.~\ref{fig:mixing_layer}, where sum node $\sumnode_1$ with $3$ children and sum node $\sumnode_2$ with $2$ children are expressed with a layer of $5$ simple sum nodes, followed by aggregated sums (sums in boxes).

The element-wise mixtures can also be implemented with a single einsum operation, which we denote as \emph{mixing layer}.
To this end, within a sum layer, let $M$ be the number of sum nodes having more than one child, and $D$ the maximal number of children.
We collect the $K$-dimensional probability vectors computed by the first (simple) sum layer in a $D \times M \times K$ tensor, where the $D$-axis is zero-padded for sum nodes with less than $D$ children.
The mixing layer computes then a convex combination over the first dimension of this tensor.

Constructing this tensor involves some copy overhead, and the mixing layer also wastes some computation due to zero padding.
However, using sum nodes with multiple children allows a much wider range of PC structures than e.g.~random binary trees, which were the only structure considered in \cite{Peharz2019}.

For sum layers which originally contain only simple sums, the construction of the mixing layer is skipped.

\section{Inference Time Comparison}

Section 4.1 in the main paper compared training time and memory consumption for EiNets, 
LibSPN \cite{Pronobis2017} and SPFlow \cite{Molina2019}, showing that EiNets scale much more gracefully than its competitors.
The same holds true for \emph{inference} time.
Fig.~\ref{fig:efficiency_comparison_supp} shows the results corresponding to Fig.~3 in the main paper, but for inference time per sample rather than training time per epoch.
Inference was done for a batch of $100$ test samples for each model, i.e.~the displayed inference time is $1/100$ of the evaluation time for the whole batch.
Again, we see significant speedups for EiNets, of up to three orders of magnitude (for maximal depth and EiNet vs.~SPFlow).

\section{On Using the Mean When Generating Images (Addendum, October 2025)}
\label{sec:addendum}

In Figure~\ref{fig:celeba}, when producing (conditional or unconditional) samples, we ``turned off the noise'' of the Gaussian leaves.
Specifically, we followed the standard top-down sampling procedure for probabilistic circuits (PCs): sum nodes (categorical distributions) are sampled hierarchically, thereby selecting a set of leaves whose scopes form a partition of all modeled pixels.
To generate an exact sample from $\PC(\mathbf{x})$, one would sample each leaf distribution and concatenate the resulting pixel values into an image.
However, instead of \emph{sampling} from the Gaussian leaves, we took their \emph{means} and concatenated these to form the final image.

This procedure is directly analogous to the common practice in variational autoencoders (VAEs), where the decoder noise is typically turned off during visualization.
Both PCs and VAEs are hierarchical latent-variable models,
\begin{align}
\mathbf{z} &\sim p(\mathbf{z}) \\
\mathbf{x} &\sim p(\mathbf{x} \mid \mathbf{z})
\end{align}
where $\mathbf{z}$ is usually continuous (Gaussian) in VAEs and discrete (associated with sum nodes) in PCs, while $p(\mathbf{x} \mid \mathbf{z})$ corresponds to the decoder in VAEs and to the hierarchical selection process in PCs, respectively.

The reason for turning off the decoder (or leaf) noise in both models is that, under the assumption of pixelwise independence in $p(\mathbf{x} \mid \mathbf{z})$, \textit{sampling would merely add independent Gaussian noise to the generated image.}
When the goal is generative modeling rather than probabilistic reasoning, this practice is justifiable, as adding Gaussian noise only degrades visual quality.

This procedure is seldom mentioned explicitly in the VAE literature, but it is common in most implementations.\footnote{See, for example, \url{https://github.com/pytorch/examples/blob/main/vae/main.py} and \url{https://keras.io/examples/generative/vae/}.}
An explicit reference to this practice can be found on Wikipedia (retrieved October 14, 2025):
\begin{quote}
``Thus, the encoder maps each point (such as an image) from a large complex dataset into a distribution within the latent space, rather than to a single point in that space. The decoder has the opposite function, which is to map from the latent space to the input space, again according to a distribution (although in practice, noise is rarely added during the decoding stage).''
\end{quote}


\bibliographystyle{icml2020}

\end{document}